\documentclass{article}
\DeclareUnicodeCharacter{FF0C}{,}

\usepackage{arxiv}
\usepackage[square,numbers]{natbib}
\usepackage[utf8]{inputenc} 
\usepackage[T1]{fontenc}    
\usepackage{hyperref}       
\usepackage{url}            
\usepackage{booktabs}       
\usepackage{amsfonts}       
\usepackage{nicefrac}       
\usepackage{microtype}      
\usepackage{lipsum}		
\usepackage{graphicx}
\usepackage{threeparttable}
\usepackage{pdflscape}
\usepackage{doi}
\usepackage{verbatim}
\usepackage{mdframed}
\usepackage{fvextra}
\usepackage[english]{babel} 
\usepackage{csquotes}
\usepackage{tabularx}
\usepackage{pifont}
\usepackage{graphicx}
\usepackage{subcaption}
\usepackage{amsmath}
\usepackage{mdframed}
\usepackage{longtable}
\newcolumntype{P}[1]{>{\raggedright\arraybackslash}p{#1}}
\usepackage{comment}
\usepackage{csquotes}
\usepackage[table]{xcolor}
\usepackage{multirow}
\usepackage{array}
\usepackage{svg}
\usepackage{diagbox}
\usepackage[normalem]{ulem} 

\usepackage{listings}
\usepackage{fancyvrb}
\usepackage{minted} 

\usepackage{soul}
\usepackage{xcolor}

\usepackage{fancyvrb}

\title{Automated Risk-of-Bias Assessment of Randomized Controlled Trials: A First Look at a GEPA-trained Programmatic Prompting Framework}

\author{ 
	Lingbo Li \thanks{Corresponding author: L.Li5@massey.ac.nz} \\
	School of Mathematical and Computational Sciences\\
	Massey University\\
	Auckland, New Zealand \\
	\And
	\href{https://orcid.org/0000-0002-9124-2536}{\includegraphics[scale=0.06]{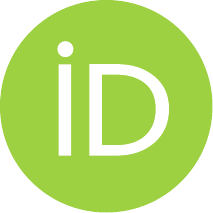}\hspace{1mm}Anuradha Mathrani} \\
	School of Mathematical and Computational Sciences\\
	Massey University\\
   Auckland, New Zealand\\
  	\And
\href{https://orcid.org/0000-0001-9416-1435}{\includegraphics[scale=0.06]{orcid.pdf}\hspace{1mm}Teo ~Susnjak} \\
	School of Mathematical and Computational Sciences\\
	Massey University\\
	Auckland, New Zealand \\
}

\renewcommand{\shorttitle}{...}

\begin{document}
\maketitle

\begin{abstract}
Assessing risk of bias (RoB) in randomized controlled trials is essential for trustworthy evidence synthesis, but the process is resource-intensive and prone to variability across reviewers. Large language models (LLMs) offer a route to automation, but existing methods rely on manually engineered prompts that are difficult to reproduce, generalize, or evaluate. This study introduces a programmable RoB assessment pipeline that replaces ad-hoc prompt design with structured, code-based optimization using DSPy and its GEPA module. GEPA refines LLM reasoning through Pareto-guided search and produces inspectable execution traces, enabling transparent replication of every step in the optimization process. We evaluated the method on 100 RCTs from published meta-analyses across seven RoB domains. GEPA-generated prompts were applied to both open-weight models (Mistral Small 3.1 with GPT-oss-20b) and commercial models (GPT-5 Nano and GPT-5 Mini). In domains with clearer methodological reporting, such as Random Sequence Generation, GEPA-generated prompts performed best, with similar results for Allocation Concealment and Blinding of Participants, while the commercial model performed slightly better overall. We also compared GEPA with three manually designed prompts using Claude 3.5 Sonnet. GEPA achieved the highest overall accuracy and improved performance by 30\%–40\% in Random Sequence Generation and Selective Reporting, and showed generally comparable, competitively aligned performance in the other domains relative to manual prompts. These findings suggest that GEPA can produce consistent and reproducible prompts for RoB assessment, supporting the structured and principled use of LLMs in evidence synthesis.
\end{abstract}


\section{Introduction}
Meta-analyses serve as cornerstone methodologies for synthesizing clinical evidence \cite{cooper_research_2017,deeks2019meta}. However, the credibility of such evidence syntheses fundamentally depends on rigorous risk of bias (RoB) assessments to evaluate the methodological quality of included studies \cite{Ma2020,Kolaski2023}. This challenge is particularly relevant for randomized controlled trials (RCTs), which constitute the primary evidence source for intervention effectiveness and require specialized assessment frameworks \cite{Higgins2024Handbook}. Currently, these assessments rely predominantly on manual evaluation processes that, while methodologically sound, face significant scalability challenges \cite{Marshall2015,Jardim2022,TOMLINSON2024}. Manual RoB assessment is inherently time-intensive, requiring expert reviewers to extract and evaluate methodological details from each trial report \cite{Marshall2015,borah_analysis_2017,higgins_cochrane_2019, TOMLINSON2024,Jardim2022}. Moreover, the process introduces unavoidable subjectivity, as different reviewers may interpret identical methodological descriptions differently, leading to inter-rater variability that can compromise the reliability of evidence syntheses\cite{Minozzi2019, Jeyaraman2020, Jyoti2021,Kaiser2023}.

Large language models (LLMs) have opened new opportunities for automating evidence synthesis \cite{li2025transform}. These models demonstrate strong capabilities in processing unstructured clinical text \cite{schmidt_exploring_2024}, understanding contextual dependencies\cite{kartchner_zero-shot_2023,shah-mohammadi_large_2024,yun_automatically_2024}, and adapting to specialized domains \cite{issaiy_methodological_2024,wei_chat2brain_2023}, making them promising tools for RoB assessment. Several studies have explored the application of LLMs for automating RoB assessment in systematic reviews and meta-analysis. Lai et al.\cite{lai_assessing_2024} applied GPT-series models and Claude to evaluate RoB in randomized trials using a structured prompt developed by experts, achieving high accuracy and substantial agreement with human reviewers. However, the study was limited by a small sample size, reliance on handcrafted prompts, and restricted validation within specific review domains, leaving its generalizability uncertain. In early 2025, Eisele-metzger et al.\cite{eisele-metzger_exploring_2025} evaluated Claude 2 using the RoB 2 tool and reported only slight-to-fair agreement with human assessments, emphasising inconsistencies in LLM performance across tools and datasets. Lai et al.\cite{lai_language_2025} further extended LLM applications to complementary medicine, demonstrating that LLM-assisted approaches achieved up to 97.9\% accuracy and markedly reduced assessment time, yet the study remained limited to complementary medicine RCTs and required expert-designed prompts for reliable performance. Most recently, Xai et al.\cite{xia_assessing_2025} applied GPT-4o, Moonshot-v1-128k, and DeepSeek-V3 to cohort studies, finding comparable accuracy (80.8–83.3\%) and substantial efficiency gains, though manual prompt construction and human oversight were still required. Across these studies, despite encouraging progress, LLM-based systems remained dependent on brittle, handcrafted prompts which exhibit high variability in outputs\cite{susnjak2025compiling}. They also possess limited-domain validation, restricting their interpretability and reproducibility required for transparent science, as well as scalability across diverse review contexts.  

In this study, we explored an alternative approach to LLM-driven RoB assessment that emphasizes transparency, while offering modularity and end-to-end integration. Rather than relying on static prompt engineering, we leveraged the DSPy \cite{khattab2023dspy} framework (a Python framework that lets users build and control LLM reasoning steps through clear, modular code rather than opaque prompts) and its GEPA \cite{agrawal2025gepa} module (Genetic-Pareto, a prompt optimization algorithm that uses natural language reflection and Pareto-based search to iteratively refine LLM reasoning) to encode domain-specific decision logic in a structured, programmable format that enhances both interpretability and maintainability. Unlike prior approaches, this framework supports few-shot learning while explicitly incorporating systematic validation, and cross-task reuse, which removes manual prompt tuning. We focus specifically on the original Cochrane RoB 1 \cite{Higgins2011RoB1} tool, capitalizing on its deterministic structure and explicit criteria that render it particularly amenable to transparent automation.

Therefore, we aim to answer a fundamental question: Can a prompt-optimization based framework like GEPA effectively automate Cochrane RoB 1 assessments for RCTs while maintaining alignment with expert human judgments? To investigate this question, we applied our methodology to a dataset of 100 RCTs that had been previously assessed by systematic reviewers in published meta-analyses, enabling direct comparison between automated model outputs and established expert evaluations across all RoB domains. This study presents an investigation to assess the feasibility of using structured, programmable LLM pipelines for evidence synthesis. Our research makes four key contributions to the field of automated RoB assessment. First, to our knowledge, this is the first empirical study to implements the prompt optimization approach to systematic reviews and meta-analysis as proposed by Susnjak \cite{susnjak2025compiling}, with an application to RoB. 
Second, we quantitatively validate GEPA-optimized prompts against expert annotations from real-world meta-analysis. Third, we evaluate both open-weight and commercial LLMs within the DSPy–GEPA workflow, highlighting trade-offs between transparency, accessibility, and performance. Fourth, we compare GEPA-generated prompts with handcrafted human prompts, showing that GEPA achieves comparable accuracy while offering substantially greater robustness and reproducibility. These contributions show that programmatic prompt optimization can meaningfully reduce reliance on manual prompt design, enhance reproducibility, and provide a scalable, transparent foundation for integrating LLMs into evidence-synthesis workflows. The modular, code-based structure of DSPy–GEPA supports adaptation across clinical domains and review settings, offering a practical pathway toward sustainable, human-oversight-compatible automation in meta-analysis.

\section{Methodology}
\begin{figure}[htbp]
    \centering
    \includegraphics[width=\textwidth]{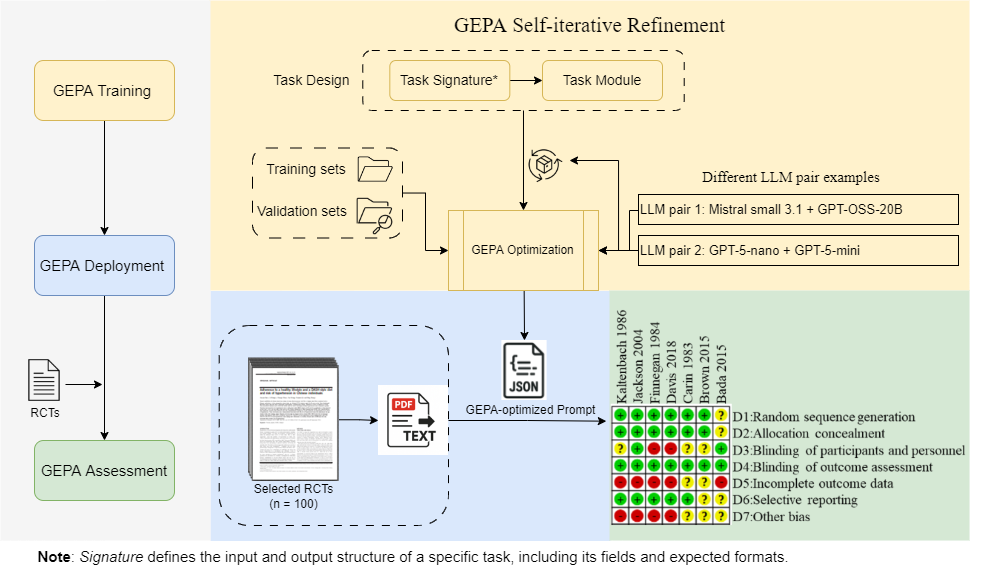}
    \caption{GEPA-based automated RoB assessment workflow}
    \label{fig:workflow}
\end{figure}

Prompt-based automation of RoB assessment currently relies on extensive manual prompt engineering and iterative refinement, often leading to inconsistent and non-reproducible outcomes\cite{suster2024}. Instead of manually crafting prompts, we used DSPy to programmatically construct and manage them through the GEPA reasoning module. This pipeline replaces manual prompt engineering with a programmatically trained process that learns and manages prompts through declarative, modular components, enabling transparent and reproducible evaluation (see Figure \ref{fig:workflow}). The following subsections describe the structure and implementation of this pipeline.

\subsection{Study Design}
We conceptualized automated RoB assessment as a structured, multi-domain decision-making process, consistent with the Cochrane RoB 1 framework. The study aimed to systematically generate domain-specific bias judgments, classified as Low, High, or Unclear risk, for each RCT across seven methodological domains: random sequence generation (whether randomization was truly random and clearly described; requires an appropriate random method), allocation concealment (whether group assignments were hidden before enrollment; requires a mechanism preventing foreknowledge of allocation), blinding of participants and personnel (whether awareness of treatment could influence behavior or care; requires masking when feasible), blinding of outcome assessors (whether knowledge of allocation could affect outcome measurement; requires blinded or independent assessors), incomplete outcome data (whether missing or excluded data were handled properly; requires balanced and transparent reporting), selective reporting (whether all pre-specified outcomes were reported; requires full outcome disclosure), and other bias (any remaining concerns such as early stopping or baseline differences; requires justification of trial conduct and design integrity). Detailed criteria for Low, High, and Unclear risk judgments in each domain are provided in Appendix Table \ref{tab:rob_criteria}.

For this study, the test dataset comprised RCTs from ten meta-analyses, randomly selected from Q1-ranked medical journals spanning diverse clinical fields, including cardiovascular disease, metabolic disorders, women’s health, and pediatrics. In total, 111 RCTs were identified; after excluding non-English trials, 100 were randomly selected for inclusion, with their distribution across clinical domains and meta-analyses detailed in Table \ref{tab:distribution}. Full-text RCTs constituted the textual corpus for automated RoB evaluation, with author-reported RoB 1 judgments from the original meta-analyses acting as the initial human reference standard (hereafter referred to as the \textit{Gold Labels}). This approach prioritized a diverse and neutral evaluation corpus, enabling comprehensive benchmarking of model performance across varied reporting styles and methodological contexts. 
The automated model’s outputs were compared against Gold Labels to evaluate prediction agreement and assess performance across RoB domains.

{
\begin{table}[htbp]
\centering
\caption{Distribution of RCTs across clinical domains and meta-analyses}
\label{tab:distribution}
\begin{tabular}{lccc}
\toprule
\textbf{Clinical Domain} & \textbf{No. of Meta-analyses} & \textbf{No. of RCTs (initial)} & \textbf{No. of RCTs (included)} \\
\midrule
Cardiovascular disease\cite{guo_effects_2021,alanouti_effects_2020,lanke_evaluating_2025} & 3 & 31 & 30 \\
Metabolic disorders\cite{khalid_effect_2023,yu_effect_2025,rose_pre-procedural_2021}    & 3 & 22 & 22 \\
Women’s health \cite{li_effects_2023, wang_effectiveness_2025}        & 2 & 30 & 23 \\
Pediatrics\cite{opioid_2021, intermittent_2021}             & 2 & 28 & 25 \\
\midrule
\textbf{Total}         & \textbf{10} & \textbf{111} & \textbf{100} \\
\bottomrule
\end{tabular}
\end{table}
}

\subsection{GEPA-based Modules}
A customized GEPA implementation was developed within DSPy to enable domain-specific reasoning for automated RoB assessment. All core components, including the curated training and validation datasets, paired DSPy Signatures definitions, module configurations were specifically designed for this study. The subsequent subsections outline the operational workflow, dataset optimization and model–environment settings.

\subsubsection{Operational Workflow}
The framework processes full-text RCTs (converted from PDF to plain text using MinerU \cite{wang2024mineru}) through a two-stage reasoning pipeline. For each of the seven Cochrane RoB 1 domains, the system employs sequential reasoning guided by paired questions based on Cochrane criteria: one focusing on identifying information relevant to the methodological process (e.g., ``Extract the exact sentence(s) describing how the allocation sequence was generated.”) and the other assessing the adequacy of the reported method (e.g., ``Was the allocation sequence adequately generated?”). The reasoning process integrates textual information and evaluative inference within a coherent reasoning framework. Evidence identification operates as an internal reasoning step that contextualizes each domain-specific judgment, while the final output represents a categorical risk-of-bias classification (``low,'' ``high,'' or ``unclear”).

\subsubsection{Dataset and Optimization}
For each RoB 1 domain, a balanced dataset was constructed, consisting of 18 training and 12 validation examples evenly distributed across the three risk categories. Each example comprised a domain-specific text excerpt from an RCT, accompanied by a structured set of elements including an evidence identification question, the corresponding evidence span, an evaluative question, the annotated risk-of-bias label, and the textual justification supporting that label. This setup supported few-shot learning and iterative supervision under constrained data conditions. GEPA optimization employed the \textit{auto=``light''} configuration (approximately 428 metric calls), balancing exploration depth with computational efficiency. All optimization runs employed fixed random seeds (42) and deterministic decoding parameters (temperature = 0.0, top\_p = 1.0). Nevertheless, because GEPA employs heuristic multi-objective search, slight variations in the Pareto optimization process can produce distinct yet equally valid optima, even under identical configurations. We assessed the stability of GEPA optimization by conducting five domain-level runs under identical settings, generating distinct optimized prompts. Each prompt was then evaluated three times on the test set to measure inference consistency and performance variability.

\subsubsection{Language Models and Environment Settings}
GEPA operates within DSPy (version used was 3.0.4b1) through a dual-model setup, comprising a base model responsible for primary reasoning and a reflection model guiding optimization feedback. Two configurations were implemented to assess performance under different computational constraints: an open-weight setup (Mistral 3.1 small\cite{mistral2025small3.1} + GPT-oss-20b\cite{openai2025gptoss}) and a commercial setup (GPT-5-nano\cite{openai2025gpt5nano} + GPT-5-mini\cite{openai2025gpt5mini}). All experiments were conducted via the OpenRouter\cite{openrouter2025api} API (September–October 2025) under uniform computational settings. This setup ensured methodological consistency and enabled direct comparison between the open-source and commercial model families.  All Python scripts are available at the GitHub repository: \url{https://github.com/Ivylee6108/GEPA_RoB.git}.

\subsection{Evaluation}
We implemented a multi-dimensional evaluation framework to assess both the internal validity and external comparability of the GEPA-based RoB assessment system. The evaluation comprised two complementary components: (1) quantitative assessment of model performance and stability, and (2) comparison with manually crafted RoB assessment prompts from prior studies.

\subsubsection{Model Performance}
Model performance was evaluated by comparing each model’s RoB predictions against Gold Labels derived from published meta-analyses. For each RoB domain, we computed standard agreement metrics, including the correct assessment rate (accuracy), sensitivity, specificity, positive predictive value (PPV), and negative predictive value (NPV), each accompanied by 95\% confidence intervals. Sensitivity was defined as the proportion of instances labeled as ``low risk'' in the Gold Labels that were correctly identified by the model. Specificity captured the proportion of instances labeled as ``unclear'' or ``high risk'' that were not correctly classified as ``low risk'' by the model. PPV represented the proportion of model-assigned ``low risk'' judgments that were also labeled as ``low risk'' in the Gold Labels, whereas NPV reflected the proportion of model-assigned ``unclear'' or ``high risk'' judgments that were consistent with the reference labels. Overall consistency was further quantified using Cohen’s Kappa coefficient ($\kappa$, 95\% CI). All metrics were calculated directly from the resulting predictions.

\subsubsection{Comparison with Manually Crafted Prompt Sets}
We also compared the proposed framework's performance against three manually crafted RoB assessment prompt sets from prior literature \cite{lai_assessing_2024, lai_language_2025, eisele-metzger_exploring_2025} (denoted as A \cite{eisele-metzger_exploring_2025}, B \cite{lai_assessing_2024}, and C \cite{lai_language_2025}). Prompt A was developed under the RoB 2 tool and originally evaluated using Claude-2. Because our framework follows the RoB 1 criteria, we matched the domains for cross-version comparability by mapping A's ``random sequence generation and allocation concealment'' domain to the ROB 1 counterparts, ``randomization process'' and ``allocation concealment''. Prompt B and C were both designed with RoB 1-like criteria but evaluated using different LLMs (ChatGPT and Claude for B, Moonshot-v1-128k and Claude-3.5-Sonnet for C). Both prompts used a four-level categorical scale (``definitely yes,'' ``probably yes,'' ``probably no,'' ``definitely no''). We harmonized these ratings with the tripartite RoB 1 scheme by mapping ``definitely yes/no'' to low/high risk and ``probably yes/no'' to unclear risk. Following this normalization, we merged their blinding-related subdomains to match the seven canonical RoB 1 domains: items concerning patients and providers were aggregated as blinding of participants and personnel, while those related to data collectors, outcome assessors, and analysts were grouped under blinding of outcome assessment. Within each merged domain, we applied a hierarchical rule: any subdomain rated ``high'' yielded an overall ``high'' rating; if all were ``low,'' the domain was marked ``low''; otherwise, ``unclear.''

Although the three prior studies employed different language models, all incorporated at least one model from the Claude family, therefore, we re-evaluated prompt A–C using Claude-3.5-Sonnet \cite{claude35_sonnet_2024}. We also maintained continuity with these Claude-based evaluation settings by adopting Claude-3.5-Sonnet as the teacher model in our GEPA framework, providing a coherent and comparable assessment environment across all approaches. Each manual prompt set (A–C) was executed three times under identical conditions, and results were averaged across runs. For the GEPA-based framework, five distinct prompts were generated through the student–teacher interaction process (student: Mistral Small 3.1; teacher: Claude-3.5-Sonnet). Each generated prompt was evaluated three times using the same RCTs in Claude-3.5-Sonnet, and the results were aggregated to yield the final framework performance. Model performance was assessed using standard quantitative metrics (accuracy) computed at both the domain and overall RoB judgment levels.

\section{Results}
This section outlines the evaluation results of our GEPA-based RoB assessment pipeline. We present overall and domain-specific performance for two pair models, comparing results against Gold Labels. 

\subsection{Overall Performance}
RoB judgments generated by the two evaluated pair-models are summarized in Table \ref{tab:overall}. The commercial model pair demonstrated higher overall accuracy and domain-level consistency compared with the open-weight pair. Across most RoB domains (D1–D5), the commercial models achieved higher correct assessment rates, with relative improvements of approximately 2–4 percentage points (e.g., 0.799 vs. 0.773 in D1 and 0.671 vs. 0.654 in D5). Sensitivity gains were similarly observed, indicating enhanced ability to identify ``low risk'' cases (0.773 vs. 0.745 in D1; 0.671 vs. 0.654 in D5). Specificity remained high for both systems, though the commercial pair showed slightly lower values in some domains (e.g., 0.888 vs. 0.910 in D1), suggesting a modestly more inclusive classification tendency. Both models achieved high positive predictive values (PPV > 0.9), while the commercial pair maintained marginally higher negative predictive values (NPV), such as 0.721 vs. 0.714 in D1 and 0.804 vs. 0.781 in D3, reflecting greater stability in ``non–low-risk'' assessments. Cohen's $\kappa$ values followed a similar pattern, with the commercial pair showing higher agreement with human reference judgments in most domains (e.g., $\kappa = 0.602$ vs. 0.567 in D1). Agreement was weaker in selective reporting (D6) and other bias (D7). Appendix Table \ref{tab:goldlabeldist}, the gold-standard label distribution was markedly imbalanced in several domains, particularly D6, where 89\% of studies were labeled ``low risk.'' Overall, the commercial models achieved higher and more consistent performance across all evaluated metrics, with improvements in accuracy, sensitivity, and NPV, and minimal trade-offs in specificity and PPV.

{
\begin{landscape}
\begin{table}[htbp]
\centering
\fontsize{8pt}{10pt}\selectfont
\begin{threeparttable}
\caption{Performance of different LLMs compared to the Gold Labels ($n=100$ RCTs)}
\label{tab:overall}
\begin{tabular}{lcccccc}
\toprule
\multirow{2}{*}{Model Pair \& RoB Domains} 
& \multicolumn{5}{l}{Accuracy} 
& \multicolumn{1}{l}{Consistency} \\ 
\cmidrule(lr){2-6} \cmidrule(lr){7-7}
& Correct assessment rate (95\% CI) & Sensitivity (95\% CI) & Specificity (95\% CI) & PPV (95\% CI) & NPV (95\% CI) & Cohen’s Kappa (95\% CI) \\ 
\midrule
\multicolumn{7}{l}{Mistral small 3.1 with GPT-oss-20b} \\
\hspace{1em} D1 (‘random sequence generation’)  & 0.773 (0.732, 0.813) & 0.795 (0.729, 0.861) & 0.910 (0.869, 0.952) & 0.945 (0.926, 0.963) & 0.714 (0.648, 0.779) & 0.567 (0.528, 0.605) \\
\hspace{1em} D2 (‘allocation concealment’) & 0.563 (0.509, 0.618) & 0.402 (0.310, 0.493) & 0.927 (0.912, 0.942) & 0.830 (0.814, 0.845) & 0.638 (0.606, 0.671) & 0.248 (0.187, 0.309) \\
\hspace{1em} D3 (‘blinding participants’) & 0.554 (0.466, 0.642) & 0.470 (0.314, 0.626) & 0.981 (0.963, 0.999) & 0.952 (0.911, 0.992) & 0.724 (0.666, 0.783) & 0.350 (0.267, 0.433) \\
\hspace{1em} D4 (‘blinding outcome assessment’)  & 0.395 (0.366, 0.425) & 0.317 (0.293, 0.342) & 0.945 (0.928, 0.963) & 0.878 (0.842, 0.915) & 0.531 (0.522, 0.541) & 0.162 (0.138, 0.185) \\
\hspace{1em} D5 (‘incomplete outcome data’) & 0.255 (0.199, 0.312) & 0.169 (0.092, 0.246) & 0.908 (0.830, 0.986) & 0.833 (0.763, 0.903) & 0.309 (0.293, 0.324) & 0.046 (0.025, 0.067) \\
\hspace{1em} D6 (‘selective reporting’)  & 0.473 (0.377, 0.569) & 0.478 (0.403, 0.553) & 0.437 (0.334, 0.539) & 0.870 (0.843, 0.898) & 0.095 (0.072, 0.118) & -0.029 (-0.076, 0.017) \\
\hspace{1em} D7 (‘other bias’)  & 0.377 (0.305, 0.448) & 0.201 (0.102, 0.301) & 0.913 (0.875, 0.950) & 0.720 (0.679, 0.761) & 0.515 (0.493, 0.537) & 0.080 (0.047, 0.113) \\
\midrule
\multicolumn{7}{l}{GPT-5 Nano with GPT-5 mini} \\
\hspace{1em} D1 (‘random sequence generation’)  & 0.795 (0.769, 0.821) & 0.811 (0.771, 0.852) & 0.888 (0.855, 0.920) & 0.932 (0.917, 0.946) & 0.721 (0.679, 0.763) & 0.602 (0.573, 0.631) \\
\hspace{1em} D2 (‘allocation concealment’) & 0.591 (0.556, 0.627) & 0.457 (0.409, 0.504) & 0.912 (0.903, 0.922) & 0.822 (0.809, 0.836) & 0.655 (0.637, 0.673) & 0.297 (0.252, 0.342) \\
\hspace{1em} D3 (‘blinding participants’)  & 0.591 (0.580, 0.603) & 0.675 (0.665, 0.684) & 0.984 (0.974, 0.994) & 0.969 (0.951, 0.987) & 0.807 (0.803, 0.811) & 0.390 (0.379, 0.401) \\
\hspace{1em} D4 (‘blinding outcome assessment’) & 0.465 (0.424, 0.507) & 0.513 (0.453, 0.572) & 0.803 (0.784, 0.822) & 0.761 (0.753, 0.769) & 0.576 (0.551, 0.601) & 0.191 (0.160, 0.222) \\
\hspace{1em} D5 (‘incomplete outcome data’)  & 0.304 (0.265, 0.343) & 0.194 (0.163, 0.226) & 0.959 (0.937, 0.981) & 0.922 (0.886, 0.959) & 0.327 (0.319, 0.336) & 0.107 (0.087, 0.127) \\
\hspace{1em} D6 (‘selective reporting’) & 0.389 (0.225, 0.554) & 0.387 (0.240, 0.534) & 0.467 (0.367, 0.567) & 0.841 (0.803, 0.880) & 0.087 (0.080, 0.093) & -0.044 (-0.054, -0.033) \\
\hspace{1em} D7 (‘other bias’)  & 0.300 (0.274, 0.326) & 0.135 (0.063, 0.206) & 0.893 (0.827, 0.959) & 0.595 (0.548, 0.641) & 0.488 (0.483, 0.493) & 0.017 (-0.004, 0.038) \\
\bottomrule
\end{tabular}
\begin{tablenotes}
\small
\item Abbreviations: CI, confidence interval; PPV, positive predictive value; NPV, negative predictive value.
\item Sensitivity: proportion correctly classified as ‘low risk’ by each evaluated model (as specified in the table) in relation to all ‘low risk’ judgements by the Gold Labels.
\item Specificity: proportion correctly classified as ‘unclear’ or ‘high risk’ by each evaluated model in relation to all such judgements by the Gold Labels.
\item PPV: proportion correctly classified as ‘low risk’ by each evaluated model in relation to all ‘low risk’ judgements by that model.
\item NPV: proportion correctly classified as ‘unclear’ or ‘high risk’ by each evaluated model in relation to all such judgements by that model.
\end{tablenotes}
\end{threeparttable}
\end{table}
\end{landscape}
}

\subsection{Performance Comparison with Manual Prompts}
\begin{figure}[htbp]
    \centering
    \includegraphics[width=\textwidth]{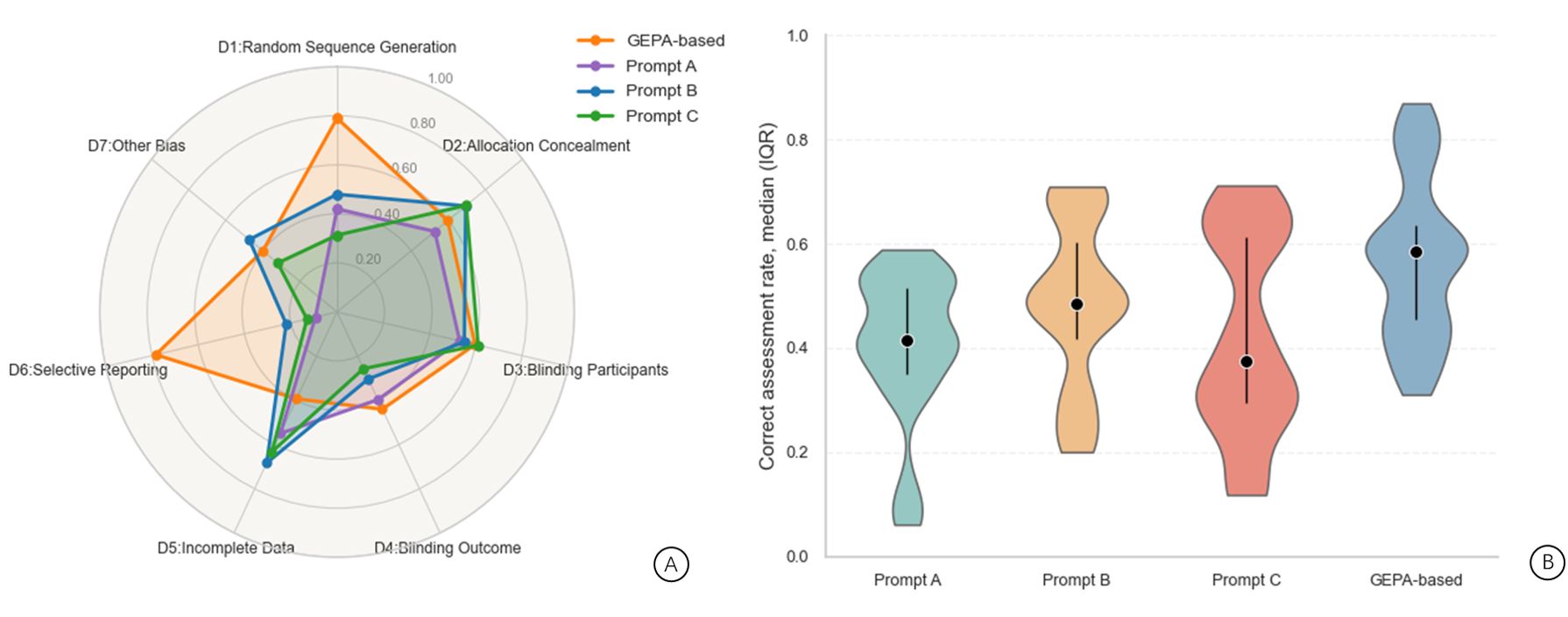}
    \caption{Performance comparison of GEPA-based and manually designed prompts. Radar plots (A) show domain-level mean accuracy across all domains, and violin plots (B) summarize overall correct assessment rates}
    \label{fig:compared}
\end{figure}

We compared the GEPA-generated prompt performance with three representative manually designed prompts (Prompt A, Prompt B, and Prompt C) to examine its generalizability and consistency (see Figure \ref{fig:compared}). The GEPA-generated prompt presents a more balanced radar shape than the manually designed prompts, suggesting steadier performance across different types of bias assessment. 
The most pronounced advantages of the GEPA-generated prompt are observed in Random Sequence Generation (mean = 0.79, 95\% CI 0.74–0.84) and Selective Reporting (mean = 0.78, 95\% CI 0.70–0.86), where its accuracy exceeds that of the best manual prompts by approximately 0.30–0.40. The higher accuracy suggests that the prompts generated by GEPA guided the LLM to make more accurate risk judgments than the manually constructed prompts. The GEPA-generated prompt also performs slightly better, though less markedly, in Blinding of Outcome Assessment (mean = 0.44, 95\% CI 0.40–0.48), showing a modest improvement over manual prompts. In Allocation Concealment (mean = 0.60, 95\% CI 0.56–0.63) and Blinding of Participants (mean = 0.60, 95\% CI 0.57–0.62), GEPA performs on par with the strongest manual prompts (mean = 0.61–0.70). For Other Bias, where only Prompts B, C, and GEPA were compared, the GEPA-generated prompt reaches a mean accuracy of 0.40 (95\% CI 0.34–0.46), comparable to Prompt B (mean = 0.47, 95\% CI 0.43–0.51) and higher than Prompt C (mean = 0.32, 95\% CI 0.29–0.35). GEPA maintains accuracy within the upper range of manual prompts. In the Incomplete Data domain, the GEPA-generated prompt performed less well than the manually designed prompts. Its mean correct assessment rate was 0.39 (95\% CI 0.33–0.45), compared with 0.55 (95\% CI 0.50–0.60) for Prompt A, 0.68 (95\% CI 0.64–0.72) for Prompt B, and 0.63 (95\% CI 0.60–0.66) for Prompt C. The lower accuracy suggests that GEPA was less effective in distinguishing between the different risk levels for incomplete data. Across all evaluations, the GEPA-generated prompts achieved the highest overall accuracy, with a median of 0.59 when averaged across the seven risk-of-bias domains. In comparison, the manually designed prompts reached medians of 0.49 for Prompt B, 0.38 for Prompt C, and 0.42 for Prompt A. Among them, Prompt A covered six domains, while Prompt B, Prompt C, and the GEPA-based approach included all seven. The higher median for the GEPA-generated prompts indicates that they enable the model to make more accurate and stable risk judgments across domains.

\subsection{Computational Efficiency and Resource Cost}
The GEPA framework demonstrated high computational efficiency and affordability across all evaluated model configurations (see Table \ref{tab:cost}). The training process consisted of two measurable components: the main generation cost, incurred by the primary LLM, and the reflection cost, generated during the self-evaluation stage by reflective LLM. Across model pairs, both components remained modest. For example, in the openweight configuration (Mistral small 3.1 with GPT-oss-20b), the main cost for each domain averaged \$0.04, with an additional reflection cost of \$0.01–\$0.02. Even when employing more capable models such as Claude 3.5-sonnet, the reflection cost remained highly manageable (\$0.2–\$0.5 per domain). Total training time ranged from approximately 22 minutes for Mistral small 3.1 with Claude 3.5-sonnet to under three hours for the GPT-5 Nano with GPT-5 mini setup. Importantly, the testing or inference cost per article was extremely low(typically below \$0.001–\$0.05), and the processing time per document remained within a few minutes, depending on the model used. These findings confirm that the GEPA procedure is transparent, quantifiable, and cost-efficient, offering reproducible and affordable time–cost trade-offs suitable for RoB synthesis applications.

{
\begin{landscape}
\begin{table}[htbp]
\centering
\fontsize{8pt}{10pt}\selectfont
\begin{threeparttable}
\caption{Computational cost and time efficiency of the GEPA-based approach}
\label{tab:cost}
\begin{tabular}{lccccc}
\toprule
\multirow{2}{*}{Model Pair \& RoB Domains} 
& \multicolumn{3}{c}{Training Procedure} 
& \multicolumn{2}{c}{Testing Procedure} \\ 
\cmidrule(lr){2-4} \cmidrule(lr){5-6}
& Main Cost (\$) (95\% CI) & Reflection Cost (\$) (95\% CI) & Time (min) (95\% CI) & Cost* (\$) (95\% CI) & Time* (min) (95\% CI) \\ 
\midrule
\multicolumn{6}{l}{Mistral small 3.1 with GPT-oss-20b} \\
\hspace{1em} D1 (‘random sequence generation’)  & 0.039 (0.031, 0.048) & 0.013 (0.006, 0.020) & 88.7 (83.8, 93.6) & \multirow{7}{*}{0.0006 (0.0005, 0.0008)} & \multirow{7}{*}{1.6 (1.6, 1.7)} \\
\hspace{1em} D2 (‘allocation concealment’) & 0.036 (0.032, 0.040) & 0.016 (0.011, 0.020) & 91.4 (86.8, 96.0) &  &\\
\hspace{1em} D3 (‘blinding participants’) & 0.040 (0.034, 0.047) & 0.028 (0.020, 0.035) & 94.2 (88.2, 100.2) &  & \\
\hspace{1em} D4 (‘blinding outcome assessment’)  & 0.036 (0.026, 0.047) & 0.010 (0.007, 0.014) & 89.8 (83.3, 96.3) &  & \\
\hspace{1em} D5 (‘incomplete outcome data’) & 0.046 (0.040, 0.052) & 0.013 (0.010, 0.016) & 92.5 (86.8, 98.2) &  & \\
\hspace{1em} D6 (‘selective reporting’)  & 0.048 (0.041, 0.056) & 0.016 (0.013, 0.018) & 87.9 (83.0, 92.8) &  & \\
\hspace{1em} D7 (‘other bias’)  & 0.041 (0.029, 0.054) & 0.014 (0.012, 0.017) & 90.6 (84.8, 96.3) &  & \\
\midrule
\multicolumn{6}{l}{GPT-5 Nano with GPT-5 mini} \\
\hspace{1em} D1 (‘random sequence generation’)  & 0.657 (0.457, 0.741) & 0.161 (0.130, 0.191) & 110.5 (95.3, 130.2) & \multirow{7}{*}{0.0014 (0.0012, 0.0019)} & \multirow{7}{*}{8.7 (8.6, 8.8)}  \\
\hspace{1em} D2 (‘allocation concealment’) & 0.550 (0.493, 0.608) & 0.111 (0.091, 0.132) & 89.7 (84.2, 95.3) &  & \\
\hspace{1em} D3 (‘blinding participants’)  & 0.726 (0.626, 0.826) & 0.150 (0.126, 0.174) & 134.1 (99.2, 168.9) &  & \\
\hspace{1em} D4 (‘blinding outcome assessment’) & 0.913 (0.631, 1.195) & 0.108 (0.086, 0.130) & 169.5 (127.3, 180.6) &  & \\
\hspace{1em} D5 (‘incomplete outcome data’)  & 0.742 (0.486, 0.997) & 0.133 (0.117, 0.149) & 133.8 (95.1, 152.4) &  & \\
\hspace{1em} D6 (‘selective reporting’) & 0.681 (0.559, 0.804) & 0.169 (0.143, 0.195) & 125.6 (104.8, 146.3) &  & \\
\hspace{1em} D7 (‘other bias’)  & 0.867 (0.596, 1.138) & 0.143 (0.114, 0.172) & 138.9 (109.1, 168.8) &  & \\
\midrule
\multicolumn{6}{l}{Mistral small 3.1 with Claude-3.5-sonnet} \\
\hspace{1em} D1 (‘randomization generation’)  & 0.030 (0.029, 0.032) & 0.254 (0.220, 0.287) & 21.3 (20.5, 22.2) & \multirow{7}{*}{0.0552 (0.0468, 0.0631)} & \multirow{7}{*}{1.7 (1.6, 1.8)}  \\
\hspace{1em} D2 (‘allocation concealment’) & 0.031 (0.030, 0.032) & 0.297 (0.264, 0.330) & 22.2 (21.2, 23.3) &  & \\
\hspace{1em} D3 (‘blinding participants’)  & 0.036 (0.035, 0.037) & 0.460 (0.429, 0.491) & 22.4 (20.7, 24.2) &  & \\
\hspace{1em} D4 (‘blinding outcome assessment’) & 0.033 (0.032, 0.034) & 0.212 (0.157, 0.267) & 18.2 (16.7, 19.7) &  & \\
\hspace{1em} D5 (‘incomplete outcome data’)  & 0.037 (0.035, 0.038) & 0.377 (0.263, 0.490) & 24.1 (21.0, 27.1) &  & \\
\hspace{1em} D6 (‘selective reporting’) & 0.035 (0.034, 0.037) & 0.402 (0.380, 0.425) & 23.1 (20.6, 25.6) &  & \\
\hspace{1em} D7 (‘other bias’)  & 0.037 (0.034, 0.040) & 0.447 (0.384, 0.510) & 25.1 (22.7, 27.4) &  & \\
\bottomrule
\end{tabular}
\begin{tablenotes}
\footnotesize
\item \textit{Note.} * Cost and time are measured per article (i.e., total cost and total processing time for one document).
\end{tablenotes}
\end{threeparttable}
\end{table}
\end{landscape}
}

\subsection{Disagreement Analysis}
We analysed disagreements across all seven RoB domains to investigate where and why GEPA-based automated judgments diverged from human ratings. Rather than assuming that such divergences necessarily reflected errors, we sought to understand how LLMs, prompted with structured reasoning templates from GEPA, may be interpreting the available evidence in ways that differ from human reviewers.
We focused on two dimensions: the direction of disagreement and the evidentiary characteristics that appeared to trigger these divergences.

\subsubsection{Overall distribution and directional tendencies}
Quantitative results showed substantial variability in GEPA–gold agreement across domains (See Table \ref{tab:overall}, Figure \ref{fig:compared}). Mean accuracies ranged from 0.30 to 0.83, and Cohen's $\kappa$ from –0.04 to 0.57, indicating that consistency was uneven and typically limited to slight or fair agreement. Only random sequence generation (D1) achieved a moderate level of agreement (mean $\kappa$ = 0.57), while most other domains fell below $\kappa$ = 0.40. An instructive outlier was D6 (selective reporting): Claude 3.5 sonnet achieved high accuracy (0.80), yet $\kappa$ was close to zero. This reflects the impact of extreme class imbalance in the gold data (89 low risk vs. 11 unclear, 0 high), which suppresses $\kappa$ despite meaningful agreement. In such cases, raw accuracy is a more valid indicator of alignment. 

Despite this variability, the direction of disagreement was remarkably consistent: models were systematically more cautious than the reference, most often downgrading low/high risk ratings to unclear. Across domains, models tended to require explicit procedural detail before assigning a low/high risk label, even in cases where human reviewers accepted adequacy based on conventional phrasing or contextual inference. Rather than indicating random inconsistency or optimism, these divergences point to a prompt-conditioned inferential style grounded in textual literalism. This ``conservatism bias'' was particularly salient when reports used formulaic expressions (e.g., ``participants were randomized,'' ``sealed envelopes were used'') without elaborating operational safeguards. In such cases, models treated absent detail as residual uncertainty, whereas human raters inferred adequacy from tacit norms.

Model scale modulated the strength of this tendency, but not the direction. Larger and more instruction-tuned LLMs achieved higher agreement, particularly in domains requiring interpretive ability (e.g., selective reporting by Claude 3.5 sonnet achieve 80\% accuracy, while mistral 3.1 small and GPT-5nano only have accuracy below 50\%, strong LLM granted low risk when the reporting was stylistically polished or implied sufficiency.) Yet even the strongest models retained the core conservatism shaped by GEPA prompts, hesitating to assign low/high risk without direct textual confirmation. This suggests that improved alignment arose not from optimism, but from smarter caution.

\subsubsection{Domain-specific disagreement illustrations}
While the overall direction of disagreement revealed a prompt-driven conservatism in model behavior, its manifestation varied across domains. This section presents illustrative examples from each domain to highlight the types of textual cues that most frequently triggered divergences, and how the LLMs' interpretive strategies conflicted with Gold Labels. These examples reflect the typical failure modes or thresholds of caution observed in that domain under GEPA-prompted inference. We include a representative example from the domain of ``allocation concealment'' in Table \ref{tab:disagree_example_main} to concretely illustrate the how LLMs demand for explicit textual confirmation led to a downgrade relative to human judgment. Additional examples across all domains are provided in Appendix Table \ref{tab:disagree_example_all}.

{
\begin{landscape}
\begin{table}[htbp]
\centering
\caption{Representative disagreement in the domain of allocation concealment}
\label{tab:disagree_example_main}
\begin{threeparttable}
\begin{tabular}{p{1.5cm} p{1cm} p{1cm} p{6cm} p{5cm} p{4cm}}
\toprule
\textbf{Trial ID} & \textbf{Gold} & \textbf{LLM} & \textbf{Text Snippet} & \textbf{LLM Justification} & \textbf{Commentary} \\
\midrule
\cite{Chattha2008} & Low & Unclear & ``The women who satisfied the inclusion criteria were registered in different nodal centres by using pre-labelled envelopes to avoid selection bias; roll numbers were allotted and these numbers were randomly divided into two groups using a computer-generated random number table (http://www.randomizer.org) prepared for the specific number of participants available in the centre.''
 & 1. The study mentions using ``pre-labelled envelopes'' but does not provide essential details about their properties:
   - No mention if envelopes were sequentially numbered
   - No specification if envelopes were sealed
   - No indication if envelopes were opaque \newline
2. While computer-generated randomization is described, this relates to sequence generation rather than concealment \newline
3. The process of how the allocation was kept concealed from study personnel is not explicitly described \newline
4. Missing information about who controlled the envelope system and how it was implemented \newline
5. Without these specific details about the concealment process, we cannot be confident that foreknowledge of assignments was prevented & We Agree the LLM’s ``Unclear'' rating is appropriate, as the study does not provide sufficient information about the envelope concealment process (e.g., sealing, numbering, or opacity), preventing a confident low-risk assessment.\\
\bottomrule
\end{tabular}
\end{threeparttable}
\end{table}
\end{landscape}
}

\paragraph{D1: Randomization generation}
Disagreements in this domain were relatively rare, with most labels aligning between LLM predictions and the gold standard. When disagreements occurred, they typically reflected the influence of the GEPA-generated prompts, which emphasized the need for explicitly documented randomization procedures. For example, studies that merely stated ``participants were randomized'' were frequently labeled unclear by the model, even when human reviewers judged them as low risk. Since the GEPA framework was trained on annotated data to produce prompts that foreground direct evidence (e.g., ``computer-generated sequence'', ``random number table''), the resulting prompts shaped the model's interpretive stance to demand such concrete phrasing before assigning low risk. Rather than relying on conventional assumptions, the model treated vague or formulaic expressions as insufficient, reflecting a more text-bound evidentiary threshold encouraged by the GEPA-guided prompting.

\paragraph{D2: Allocation concealment}  
This domain showed a high rate of disagreement, primarily reflecting LLMs' insistence—guided by GEPA prompts—on detailed procedural transparency. GEPA prompts highlighted key indicators such as opaque, sequentially numbered, and independently administered concealment methods. In contrast, many trial reports used generic phrases like ``envelopes were used'' without specifying safeguards. The LLMs, following GEPA's stricter evidentiary framing, typically assigned unclear unless all key qualifiers were present.

\paragraph{D3: Blinding participants}  
Disagreements in this domain were frequent and predominantly reflected the conservative evidentiary stance encouraged by the GEPA-generated prompts. In trials involving physical, behavioral, or lifestyle interventions, the models rarely assigned low risk unless explicit descriptions of blinding procedures were present. GEPA's structured prompting emphasized the need to verify whether participants and personnel were blinded and whether the lack of blinding could plausibly influence performance-related outcomes. As a result, when reports did not clearly state whether blinding was implemented, or when blinding was inherently infeasible due to the nature of the intervention, the LLMs defaulted to high or unclear risk.

\paragraph{D4: Blinding outcome assessment}  
Disagreements in this domain were common and primarily stemmed from the models' insistence on explicit evidence of assessor blinding. Guided by GEPA-derived prompts, the LLMs required concrete details, such as who conducted the outcome assessments, whether they were blinded, and how blinding was implemented. Human reviewers, by contrast, frequently assigned low risk when measurement tools were objective or standardized, inferring adequate protection against detection bias. This divergence reflects GEPA's conservative evidentiary stance: lacking direct textual confirmation, the model defaulted to unclear, prioritizing procedural transparency over contextual inference.

\paragraph{D5: Incomplete outcome data}  
Disagreements in this domain also reflected the LLMs’ limited access to information presented in non-textual elements, such as CONSORT flow diagrams. While GEPA-trained LLMs required explicit numerical reporting (e.g., dropout counts, reasons for attrition) and detailed analytical descriptions (e.g., intention-to-treat implementation) in the text before assigning low risk, human reviewers often integrated information from figures or supplementary materials. Consequently, human raters judged some cases as low risk when visual data indicated minimal, balanced attrition, even though the textual record alone appeared incomplete. This divergence highlights how LLMs, constrained by textual input and GEPA’s rigid evidentiary thresholds, may underestimate reporting adequacy compared to human reviewers’ more holistic interpretations.

\paragraph{D6: Selective reporting}  
Disagreements in this domain were shaped by LLMs' differing thresholds for what constitutes sufficient evidence of outcome reporting transparency. Despite being trained on GEPA examples where comprehensive outcome reporting sufficed for low risk, even without quoting registry contents, smaller models (e.g., Mistral, GPT-5 nano) often defaulted to unclear unless a direct comparison to a registered protocol was present. This conservatism persisted even when reports listed all primary and secondary outcomes transparently. In contrast, stronger models like Claude more accurately captured the intended standard. When trial reports mentioned a protocol or registration and provided detailed, structured outcome reporting, Claude was more likely to assign low risk, reflecting better generalization of the training signal. Divergence in this domain was therefore more about the models' varying ability to infer adequacy from indirect cues.

\paragraph{D7: Other bias}  
Disagreements in this domain were highly variable and largely stemmed from the inherently subjective nature of the ``Other bias'' category in RoB 1. Human reviewers themselves often applied this domain inconsistently, as it covers a wide array of loosely defined concerns, ranging from funding sources to adherence issues or baseline imbalances. GEPA-trained LLMs struggled to apply stable criteria, particularly when trial texts mentioned potential bias-adjacent issues (e.g., small sample sizes or recruitment concerns) without clarifying their methodological implications. In such cases, models typically defaulted to unclear, reflecting both a conservative stance and ambiguity in the task framing. Some disagreements arose from the models' heightened caution, while others reflected divergent assumptions about what constitutes a risk-relevant concern.

\section{Discussion}
In this study, we introduce a programmatic approach for optimizing prompts used in LLM, namely assisted RoB assessment. Rather than relying on manually crafted instructions, which can vary widely across different teams, GEPA formalizes prompt development as an explicit, data-driven optimization procedure. Our findings demonstrate that optimized prompts can meaningfully improve the consistency and robustness of automated RoB judgments, while also revealing important methodological considerations related to the inherent subjectivity of RoB assessment, domain-specific variability, and the continued need for human oversight.

\subsection{From Hand-Crafted to Programmatically Optimized Prompts}  
Prompt development for LLM-assisted evidence synthesis has historically relied on expert intuition and iterative trial-and-error. Hand-crafted prompts embed tacit heuristics and stylistic tendencies that differ across reviewers, leading to substantial variability and limited reproducibility. Moreover, manual prompts are inherently volatile: small changes in wording or formatting can produce disproportionately different LLM outputs, undermining the stability required for reliable RoB assessment. This fragility is particularly consequential in domains where judgments depend on contextual interpretation. As a result, handcrafted prompts constrain the transparency, portability, and standardization needed for evidence-based workflows and create persistent challenges whenever models are updated or replaced.
GEPA addresses these methodological limitations by formalizing prompt design as a structured optimization problem. Rather than depending on subjective refinement, GEPA systematically searches a broad space of candidate prompts and identifies those that satisfy predefined criteria for accuracy, faithfulness, and conciseness. Crucially, the entire optimization process produces comprehensive execution traces, including intermediate candidates, reasoned model outputs, Pareto-front evaluations, and the final compiled prompt, which can be inspected, versioned, and shared with the scientific community. This level of transparency is rarely achievable with handcrafted prompts and represents a substantial advance in the reproducibility of LLM-driven RoB assessment. Because every step of the search and selection process is logged, GEPA produces prompts that are transparent, reproducible, and model-agnostic. Our experiments show that optimized prompts transfer effectively across related models, indicating that GEPA captures task-level regularities instead of overfitting to model-specific linguistic quirks. This methodological stability enables GEPA to serve as a durable foundation for LLM-based evidence synthesis, even as LLM architectures continue to change.

\subsection{Human Oversight with Minimal Prompt Design Burden}  
Beyond methodological reproducibility, GEPA-optimized prompts inform on how human reviewers interact with LLM-assisted workflows. In conventional pipelines, a major barrier to reliable automation is the substantial manual effort required to design and validate prompts for each RoB domain. This task demands methodological expertise and can vary widely across reviewers, introducing inefficiencies and inconsistencies into the workflow. As prompts often need re-engineering when models evolve, maintaining such pipelines becomes labor-intensive and error-prone.
GEPA alleviates this burden by generating stable, criteria-oriented prompts without requiring manual engineering. With optimized prompts in place, LLMs can provide consistent, text-grounded first-pass assessments, allowing human reviewers to redirect their expertise toward higher-value activities, such as adjudicating ambiguous cases, integrating evidence across trials, and ensuring coherence with the broader literature. This redistribution of effort enhances efficiency, reduces workflow fragility, and is particularly advantageous for teams with heterogeneous training or when processing large volumes of evidence rapidly. Rather than replacing human judgment, GEPA supports a more structured, dependable human-in-the-loop process by removing the need for handcrafted prompts.

\subsection{Domain-Specific Variability}
The performance of GEPA-optimized prompts requires deeper consideration on the conduct of human RoB assessment itself. Prior works \cite{Jorgensen2016RoBTool, Minozzi2020RoB2, Konsgen2020InterReview} repeatedly shown that reviews differ in how strictly they apply RoB criteria, and RoB judgments are inherently subjective, with only moderate agreement even among trained reviewers. Many systematic review teams rely on contributors with varying levels of methodological expertise, and reviewers frequently draw on tacit knowledge, prior experience, or domain-specific expectations that are not explicitly codified in guidance documents. These factors introduce variability into human assessments that is difficult to eliminate, even with standardized tools such as the Cochrane RoB tool. As a result, the GEPA-optimized prompts reflect of a mixture of high-quality judgment, heterogeneous heuristics, and variable interpretive practices. Their judgement do not affect all RoB domains equally. Domains with clearly specified criteria, such as random sequence generation and allocation concealment, tend to elicit more consistent human assessments, providing GEPA with a stronger and more coherent supervisory signal. In contrast, domains such as other bias or incomplete outcome data involve greater interpretive flexibility, lower inter-rater reliability, and more heterogeneous reasoning. GEPA faces an intrinsically more difficult optimization problem, and performance ceilings naturally differ across domains. However, this variability also highlights an important methodological advantage of GEPA. By optimizing prompts against patterns that are stable across many annotations, GEPA distills a more standardized and reproducible interpretation of RoB criteria than can typically be achieved by individual reviewers. In contrast to handcrafted prompts, GEPA learns decision boundaries shaped by aggregate judgment behavior, smoothing over idiosyncratic tendencies. Through this process, GEPA tends to adopt a more conservative and criteria-driven stance, particularly when textual justification is incomplete. This strictness corresponds to the normative intent of RoB tools, which emphasize cautious inference when reporting is insufficient, and contrasts with the more permissive patterns sometimes observed among untrained or inconsistently trained reviewers. Therefore, within the constraints of what is documented in trial reports, GEPA provides a stable, defensible baseline that is often more consistent than typical human practice, especially in resource-limited review settings.

\subsection{Implications for Evidence Synthesis}
Adoption of GEPA has implications that extend beyond automating discrete components of RoB assessment. By replacing handcrafted prompts with a reproducible optimization procedure, GEPA offers a practical mechanism for standardizing the interface between human judgment and LLM-based reasoning. This development is particularly relevant for systematic reviews and living evidence syntheses, where variability in reviewer expertise and constraints on time and resources often limit the consistency of methodological appraisal. Equally important, GEPA’s complete execution logs provide an auditable record of how reasoning strategies were derived. These traces can be packaged and shared alongside the compiled prompts, facilitating scientific replication, model-to-model comparison, and long-term reproducibility in evidence synthesis workflows. More broadly, GEPA illustrates how programmatic optimization can be used to formalize and stabilize workflows that have traditionally relied on tacit knowledge or iterative trial-and-error. Embedding such standardized components into evidence synthesis pipelines has the potential to enhance reproducibility, reduce reviewer workload, and support scalable review processes as evidence bases continue to expand. While human oversight remains essential, GEPA demonstrates a pathway toward more structured, transparent, and sustainable integration of LLMs in evidence synthesis practice.

\section{Conclusion, Limitations, and Future Research}

This study introduced a programmatic and fully auditable approach to automating risk-of-bias assessment, replacing volatile manually crafted prompts with a transparent and optimizable computational process. The GEPA-based optimization pipeline systematically learns and refines reasoning strategies through structured Pareto-guided search rather than ad-hoc textual instructions. The method achieved expert-level consistency in objective domains, such as random sequence generation and allocation concealment, while maintaining interpretability, transparency, and reproducibility. Importantly, every step of the optimization procedure generates inspectable execution traces that can be packaged and shared to support full scientific replication. In more subjective domains, including blinding outcome assessment and other bias, its cautious performance highlights the continued need for contextual reasoning and human oversight.

Despite these contributions, several limitations warrant attention and provide directions for future research. First, our framework is built around the RoB 1 tool, which relies on domain-level judgments and does not incorporate the signaling-question–based structure introduced in RoB 2. Extending GEPA to accommodate multi-step reasoning chains and conditional decision rules would enhance its applicability and alignment with current methodological standards. Second, our preprocessing relied on PDF-to-text conversion, resulting in the loss of figures and flow diagrams that may contain relevant methodological information. Integrating multimodal inputs or structured metadata may help mitigate this limitation. Third, the dataset used for optimization was selected to balance coverage and feasibility, but the optimal scope and characteristics of such datasets remain unclear. Variability in human RoB judgments, particularly in domains with lower inter-rater agreement, also limits the achievable ceiling for fully automated systems. Future work should examine how larger, more diverse, and more consistently annotated datasets could improve stability and generalizability.

Beyond these methodological considerations, GEPA illustrates a broader opportunity for extending programmatic optimization within evidence synthesis. By replacing ad-hoc prompt design with a principled optimization pipeline, the framework demonstrates how interpretive reasoning tasks can be formalized, audited, and scaled. Although the present study focuses on RoB assessment, the same principles could be applied to other components of meta-analytic workflows, including literature retrieval, data extraction, and evidence aggregation, where reproducibility, traceability, and scalable human oversight are essential.

\bibliographystyle{unsrtnat}
\bibliography{references}

\appendix  
\section{Criteria for Low, High, and Unclear Risk Judgments in Each RoB Domain}

{
\begin{landscape}
\begin{table}[htbp]
\centering
\caption{Criteria for Low, High, and Unclear Risk Judgments in Each RoB Domain}
\label{tab:rob_criteria}
\begin{tabular}{p{3cm} p{5cm} p{5cm} p{5cm}}
\toprule
\textbf{Domain} & \textbf{Low Risk} & \textbf{Unclear Risk} & \textbf{High Risk} \\
\midrule
Sequence generation & Randomization method clearly described and appropriate (e.g., computer-generated, random number table). & Randomization mentioned but method not described. & Non-random or predictable method (e.g., alternation, date of birth). \\
Allocation concealment & Concealment mechanism prevents foreknowledge (e.g., sealed opaque envelopes, central allocation). & Insufficient details on concealment method. & Allocation predictable or not concealed. \\
Blinding of participants and personnel & Blinding implemented and likely effective. & Blinding unclear or partial. & No blinding, and outcomes likely influenced by knowledge of intervention. \\
Blinding of outcome assessment & Outcome assessors blinded and unlikely to be influenced. & Blinding not reported or uncertain. & Assessors not blinded and outcomes subjective or easily biased. \\
Incomplete outcome data & Few or balanced dropouts; reasons explained; appropriate handling (e.g., ITT analysis). & Incomplete data or handling not described. & High or unequal attrition without adequate explanation or handling. \\
Selective reporting & All prespecified outcomes reported as planned. & Unclear whether all outcomes reported. & Evidence of omitted or selectively reported outcomes. \\
Other bias & No additional concerns. & Possible but unverified sources of bias. & Clear evidence of other problems (e.g., early stopping, baseline imbalance). \\
\bottomrule
\end{tabular}
\end{table}
\end{landscape}
}

\section{Distribution of gold-standard RoB judgments across seven domains}
{
\begin{table}[ht]
\centering
\caption{Distribution of gold-standard RoB judgments across seven domains}
\label{tab:goldlabeldist}
\begin{tabular}{lccc}
\hline
\textbf{Domain} & \textbf{Low Risk} & \textbf{Unclear} & \textbf{High Risk} \\
\hline
D1 Random sequence generation & 65 & 28 & 7 \\
D2 Allocation concealment & 47 & 42 & 11 \\
D3 Blinding of participants & 42 & 28 & 30 \\
D4 Blinding of outcome assessment & 55 & 21 & 24 \\
D5 Incomplete outcome data & 71 & 13 & 16 \\
D6 Selective reporting & 89 & 11 & 0 \\
D7 Other bias & 52 & 32 & 16 \\
\hline
\end{tabular}
\end{table}
}

\section{Illustrative disagreements across all RoB domains}

{
\begin{landscape}
\fontsize{8pt}{10pt}\selectfont
\begin{longtable}{p{0.6cm} p{1.7cm} p{0.6cm} p{1cm} p{7cm} p{5cm} p{4cm}}
\caption{Representative disagreement examples across all RoB domains}
\label{tab:disagree_example_all} \\
\toprule
Domain & Trial ID & Gold & LLM & Text Snippet & LLM Justification & Commentary \\
\midrule
\endfirsthead

\caption*{(Continued) Representative disagreement examples across all RoB domains} \\
\toprule
Domain & Trial ID & Gold & LLM & Text Snippet & LLM Justification & Commentary \\
\midrule
\endhead

\midrule
\multicolumn{6}{r}{Continued on next page} \\
\endfoot

\bottomrule
\endlastfoot
D1 &\cite{Bada2015} & Low & Unclear & ``When the FS met criteria for pharmacologic treatment and after consent, the infant received the study drug based on the randomization schedule (4 per block), available only through the investigational drug unit of the Department of Pharmacy."& 1. The evidence indicates use of a block randomization approach (4 per block), which is a valid randomization method in principle
2. However, crucial details are missing:
   - The method used to generate the randomization schedule is not specified
   - No information about who generated the sequence
   - No description of the actual sequence generation process
3. While the involvement of the pharmacy department suggests proper implementation, this relates more to allocation concealment than sequence generation
4. The lack of specific information about how the randomization schedule was created prevents a ``Low" risk assessment
5. There are no clear signs of systematic bias that would warrant a ``High" risk assessment
Therefore, due to insufficient information about the sequence generation process, an ``Unclear" risk assessment is most appropriate.
&
We agree the LLM’s “Unclear” rating is appropriate, as the study does not describe the method used to generate the randomization sequence, despite mentioning block randomization.

\\
D2 &\cite{Brahe2015} & Low & Unclear & ``Enrolment of participants was performed continuously by the same two researchers, and randomisation was carried out continuously using pre-prepared opaque sealed envelopes with an equal allocation ratio of 1:1:1."
&1. The study uses opaque sealed envelopes, which is a potential method for adequate concealment
2. However, two critical aspects raise concerns:
   - The envelopes are not described as sequentially numbered, which is essential for preventing selection bias
   - The same researchers handled both enrollment and randomization, which could potentially compromise the concealment process
3. While some concealment measures were in place (opaque sealed envelopes), the description lacks sufficient detail to ensure the process was truly secure
4. Without confirmation of sequential numbering and separation of duties, we cannot be certain that foreknowledge of assignments was prevented
5. Therefore, while not clearly high risk, there is insufficient information to confirm adequate concealment, leading to an unclear risk assessment
&
We agree the LLM’s “Unclear” rating is appropriate, as the study lacks key details about allocation concealment, including whether envelopes were sequentially numbered and whether roles in enrollment and assignment were separated.

\\
D3 &\cite{Innes2012} & Low & High & ``All assessments were conducted by experienced General Clinical Research Center and university clinical laboratory staff blinded to participant treatment assignment. Following enrollment, screening, and baseline assessment, eligible participants were randomized, using a computer-generated randomization list provided by a statistician not associated with the study, to receive either a gentle 8-week yoga program (n = 38) or an 8-week educational film program (n = 37). Each participant was administered her treatment assignment sequentially in order of enrollment, via coded opaque envelopes containing general study instructional materials, a welcome letter, and information pertinent to the yoga or educational film program. Each group (yoga and film intervention) attended a 90-minute class twice weekly for 8 weeks. Classes for the two programs were scheduled at the same times and in similar settings. Film group participants also received the yoga home practice materials, as well as coupons to attend local Iyengar yoga classes. Similarly, yoga group participants were also offered the opportunity to view the educational films following completion of the final assessment. Participants randomized to the yoga program completed a gentle Iyengar yoga program specifically designed for older, sedentary adults. In addition to attending classes, participants were asked to complete at least 30 minutes of home practice on nonclass days, with the aid of a DVD and a booklet illustrating the yoga home practice routines. This standardized educational film program, described in detail elsewhere, was designed to be easily replicable and to ensure comparability of staff attention, class time, and social interaction. Each class began with a brief meet and greet period, followed by viewing of an educational film chosen for its relevance and interest to our study population, then concluded with a 10-15 minute group discussion facilitated by a health professional with expertise in complementary and alternative therapies and women's health. To increase credibility and reduce participant bias, the educational film program was presented to potential participants as an informative, engaging, and relaxing alternative to the yoga classes. Participants underwent a comprehensive assessment at baseline and following the 8-week intervention period."
&1. The study design inherently precludes effective blinding of participants and personnel:
- Participants actively engaged in either yoga practice or watching films, making it impossible for them not to know their assignment
- Yoga instructors and film program facilitators necessarily knew which intervention they were delivering
- The interventions were fundamentally different in nature (physical activity vs. passive viewing)

2. While the study implemented some measures to minimize bias:
- Assessment staff were blinded to treatment assignment
- The educational film program was presented as a credible alternative
- Similar class times, settings, and attention levels were maintained
These measures help reduce some forms of bias but do not achieve true blinding of participants and personnel.

3. The inability to blind participants and personnel could influence outcomes, particularly for self-reported measures like sleep quality, stress, mood, and anxiety, which are susceptible to performance bias.
& We agree the LLM’s “High” rating is appropriate, as blinding of participants and personnel was not feasible given the nature of the interventions, increasing the risk of performance bias for subjective outcomes.
\\
D4 &\cite{Jayabharathi2014} & Unclear & High & ``QoL was assessed by the investigator using the same questionnaire at weeks 6, 12, and 18, and reinforcement sessions were implemented following each test. The investigator remained with the women while they completed the questionnaire and answered any questions.

The questionnaire used in this study comprised two sections. Section 1 pertained to information regarding demographic data, including age, religion, marital status, type of family, availability of a support system, age of menarche, parity, and duration of attainment of menopause. Section 2 consisted of the standardized WHO QoL-BREF scale to assess QoL in menopausal women. The Tamil version of the WHO QoL-BREF scale was used without any modification. Formal permission was obtained from the WHO to use this scale in the present study. The scale contains 26 items under four domains, ie, a physical domain, a psychological domain, a social domain, and an environment domain. Of the 26 items, 23 are positively scored and three (items 3, 4 and 26) are negatively scored on the scale. The first two items are examined separately: item 1 asks about the subject's overall perception of QoL and item 2 asks about the subject's overall perception of health in menopausal women (satisfaction of health status). The other 24 items are included in the domains. Raw scores are transformed into 100. Domain scores are scaled in a positive direction (higher scores denoting better QoL). The mean score for items within each domain is used to calculate the domain score. Mean scores are then multiplied by 4 in order to derive the domain scores and subsequently transformed to a 0-100 scale using the formula given by the WHO. Domain-wise interpretation of the data was done using various statistical methods."
&
1. The evidence indicates that the investigator was directly involved in the assessment process, remaining with participants during questionnaire completion
2. There is no mention of any blinding procedures or attempts to prevent the investigator from knowing group assignments
3. While a standardized questionnaire (WHO QoL-BREF) was used, which provides some structure to the assessment:
   - The direct interaction between investigator and participants during assessment
   - The investigator's role in answering questions during completion
   - The lack of explicit blinding procedures
All suggest that outcome assessors likely knew participant group assignments
4. The presence of reinforcement sessions following each test further indicates ongoing interaction between investigators and participants, increasing the likelihood that group assignments were known
5. This setup presents a high risk of bias as the outcome assessor's knowledge of group assignments could influence how they interact with participants or interpret responses
&
We agree the LLM’s “High” rating is appropriate, as the investigator conducting the assessments was not blinded and interacted directly with participants during questionnaire completion, creating substantial potential for detection bias.

\\
D5 &\cite{Zou2017} & Low & Unclear &
"Sample size calculation was based on systolic blood pressure reduction, reported in DASH (15), DASH-sodium (14), and PREMIER (13). With a power of 80\% and two-tailed significance level of 5\%, to detect a 3 mm Hg difference on systolic blood pressure measurement between study groups (within-group standard deviation 10 mm Hg), a total sample size of 350 (175 per group) would be required. To account for an estimated 10\% loss to follow-up, a sample size of 386 (193 per group) would be needed, but because this was a pilot trial, only 60 participants (30 per group) were recruited."

"In total, 57 (95\%) participants attended the 8 weeks post-randomization follow-up session."

"All participants were analyzed in the group to which they were assigned following the intention to treat principle."
&
While some aspects suggest good handling of missing data:

1. Very low overall dropout rate (5\%, only 3 participants)  
2. Use of intention-to-treat analysis  
3. High retention rate (95\% completion)  

Several missing elements:  

- No information about group distribution of the 3 dropouts  
- No documentation of reasons  
- No details about how missing data was handled  
- Cannot assess if missingness related to outcomes  

Despite the low dropout rate, the missing information prevents a confident Low risk assessment. The ITT use is positive, but insufficient to confirm robustness.
&We disagree with the LLM’s “Unclear” rating. Although the text lacks detail, the flow diagram (Figure 1) provides sufficient information on dropout distribution and reasons, supporting a “Low” risk assessment for missing outcome data.
\\
D6 &\cite{Lee2012} & Low & Unclear &
"The variables of body composition, serum lipids, MS factors, and energy metabolism were measured in all women before the start of the training program and at the end of the 16-week yoga exercise program.
The yoga exercise group and control group at baseline were similar in body weight, percentage of BF (\% BF), lean body mass, BMI, WC, and VFA.
After the yoga exercise program, the yoga group had significantly lower body weight, \% BF, BMI, WC, and VFA (P < 0.05) than did the control group; however, lean body mass did not change.
The yoga exercise group had significantly higher adiponectin (P < 0.001) and lower TC, systolic blood pressure (SBP), DBP, glucose, and HOMA-IR than did the control group after the training program (P < 0.05).
In a multiple linear regression analysis, serum adiponectin concentrations were significantly correlated with WC, HDL-C, DBP, and HOMA-IR in the postyoga exercise group.
The yoga exercise program induced a decrease in body weight, \% BF, BMI, WC, and VFA.
Yoga training provided continuous improvement in serum adiponectin, MS factors, insulin, and HOMA-IR.
The yoga exercise group had significantly higher adiponectin and lower blood pressure and HOMA-IR compared with the control group.
Yoga exercise significantly decreased body weight, \% BF, BMI, WC, VFA, TC, TG, LDL-C, blood pressure, insulin, glucose, and HOMA-IR.
The results of this investigation revealed a significant improvement in both SBP and DBP compared with baseline values.
Adiponectin also showed a significant positive correlation with HDL-C and significant negative correlations with DBP and HOMA-IR.
The concentration of adiponectin increased during the yoga exercise program, with a significant negative correlation being observed between adiponectin and WC.
The study was approved by the Institutional Human Research Committee.
The participants were randomized to a yoga exercise group (n = 8) or to a no exercise (control group (n = 8) for 16 weeks.
The intervention trial was designed to compare pre-exercise and postexercise intervention variables.
Changes from baseline to the end of the intervention were determined using a paired t test and an independent t test.
A power analysis was conducted using a sample size for each group of eight, obtaining an estimated power of 0.81 for a paired-sample t test."
&
The risk level is assessed as ``Unclear" for the following reasons:

1. Protocol/Registry Status:
- No mention of a pre-registered protocol or trial registry
- Cannot verify if outcomes were pre-specified as reported

2. Comprehensive Reporting Evidence:
- Clear reporting of primary outcomes (body composition, serum lipids, MS factors)
- Complete statistical analysis methods described
- Both significant and non-significant results reported
- Detailed baseline and post-intervention data provided
- Power analysis included

3. Transparency Factors:
- Results presented for all mentioned variables
- Both positive and negative findings reported
- No indication of selective reporting
- Statistical methods appropriate and clearly described

While the reporting appears thorough, without a protocol to compare against, we cannot definitively rule out selective reporting. This leads to an ``Unclear" rather than ``Low" risk assessment, despite the apparently complete reporting.
& We agree the LLM’s “Unclear” rating is appropriate. Although outcome reporting appears complete and balanced, the absence of a protocol or trial registration prevents verification that outcomes were prespecified. Thorough reporting alone is not sufficient to rule out selective reporting bias.
\\
D7 &\cite{Elavsky2007} & Unclear & High &
"Participants in the control group were found to be slightly younger than participants in the walking or yoga groups (p < .05) and participants in the yoga group had more children (p < .05) and higher income (p < .05). There were no other baseline differences among participants. Including age as a covariate in the analyses changed the pattern of results only for menopause-related QOL and menopausal symptoms. All remaining analyses are presented unadjusted."

"Dropouts differed from those who completed the study relative to baseline body mass index (dropouts 35.5 vs. 28.9, p = .001), trunk fat (dropouts 41.4\% vs. 35.5\%, p < .01), waist-to-hip ratio (dropouts .88 vs. .81, p = .001), and total body fat (dropouts 40.5\% vs. 37.3\%, p < .05)."

"Although an intent-to-treat approach was adopted for the analyses, there were some missing data across assessments due to participant refusal to answer certain questions, failure to return some questionnaires, or refusal to participate in physiological testing or body composition assessment using the DXA.""
&
The high risk assessment is justified through the following steps:

1. Baseline Imbalances:
- Significant differences existed between groups in age, number of children, and income
- While some statistical adjustments were made, they only affected specific outcomes
- Other outcomes remained unadjusted despite known baseline differences

2. Systematic Differences in Dropouts:
- Dropouts had significantly different physical characteristics
- Higher BMI, trunk fat, and body fat percentages in dropouts suggest systematic attrition
- This pattern could bias results if related to treatment effectiveness

3. Incomplete Data Collection:
- Multiple sources of missing data identified
- Participant refusal for certain assessments suggests non-random missing data
- Despite intent-to-treat approach, missing data could affect outcome validity

4. Cumulative Impact:
- The combination of baseline imbalances, systematic dropout patterns, and missing data creates multiple potential sources of bias
- These issues could systematically affect the study results in ways that statistical adjustments may not fully address
& We agree the LLM’s “High” rating is appropriate, as the study shows evidence of multiple potential sources of other bias. These include baseline imbalances in age, number of children, and income; systematic differences between dropouts and completers in key physiological variables; and non-random missing data due to participant refusal. Such issues are not addressed fully by statistical adjustments and could affect the internal validity of the findings.
\\
\end{longtable}

\end{landscape}

}

\section{Examples of GEPA-trained Prompts}
\subsection{Mistral small 3.1 with GPT-oss 20b}

\begin{lstlisting}[language=]
markdown
## Instruction for Risk‑of‑Bias Assessment (Allocation Sequence Generation)

You will be given a **question** about the adequacy of the allocation sequence in a randomized study and a short excerpt of the study’s description (the *evidence*). Your task is to produce a structured answer with four fields:

| Field | Purpose | Format |
|-------|---------|--------|
| **reasoning** | A concise, step‑by‑step explanation of how you evaluated the evidence. | Plain text, 1–3 sentences. |
| **risk_level** | The overall risk of bias judgment for the allocation sequence. | One of: `Low`, `Unclear`, `High`. |
| **justification** | A brief narrative that ties the reasoning to the risk level. | Plain text, 2–4 sentences. |
| **confidence** | Your confidence in the judgment, expressed as a decimal between 0 and 1. | Numeric value (e.g., `0.85`). |

### Domain‑Specific Knowledge

1. **Randomization**  
   - *Adequate* if the study explicitly states that participants were randomly assigned (e.g., “randomly assigned,” “randomized”) and describes a method that gives each participant an equal chance of assignment (e.g., computer‑generated random numbers, random number tables, dice rolls).  
   - *Inadequate* if assignment was based on clinician judgment, patient preference, or any deterministic rule.  
   - *Unclear* if the description is missing, vague, or does not specify how randomness was achieved.

2. **Stratification**  
   - Stratified randomization (e.g., “stratified by sex and baseline pain intensity”) is acceptable **provided** the stratification variables were used to generate separate random sequences.  
   - If stratification is mentioned but no randomization method is described, the risk remains *Unclear*.

3. **Allocation Concealment**  
   - Not required for the *sequence generation* domain, but if the evidence includes concealment details (e.g., sealed envelopes, central randomization), note it in the reasoning.  
   - Lack of concealment does **not** automatically raise the risk for sequence generation, but it may be relevant for other bias domains.

4. **Common Pitfalls**  
   - **Non‑random assignment** (e.g., “clinicians decided based on presentation”) → **High** risk.  
   - **Random method described but no details** (e.g., “randomly assigned” with no method) → **Unclear** risk.  
   - **Random method described with clear, reproducible procedure** → **Low** risk.

### Decision Rules

| Evidence | Risk Level |
|----------|------------|
| Clear random method (e.g., die roll, computer randomization) | Low |
| Random method described but lacking detail or reproducibility | Unclear |
| No random method, deterministic or clinician‑based assignment | High |
| Evidence missing or ambiguous | Unclear |

### Confidence Scoring

- **0.90–1.00**: Very high confidence (clear evidence, no ambiguity).  
- **0.70–0.89**: High confidence (minor uncertainty).  
- **0.50–0.69**: Moderate confidence (some ambiguity).  
- **<0.50**: Low confidence (significant uncertainty).

### Output Example

```yaml
reasoning: "The study reports that participants were randomly assigned using a computer‑generated sequence, with stratification by sex and baseline pain intensity."
risk_level: Low
justification: "The explicit mention of a computer‑generated random sequence and stratification indicates a robust randomization process, minimizing selection bias."
confidence: 0.92
\end{lstlisting}

\subsection{GPT-5 Nano with GPT-5 mini}
\begin{lstlisting}[language=]
Prompt:
You are to assess risk-of-bias for the ROB1 criterion "Was the allocation sequence adequately generated?" using only the provided evidence. Your output must follow the exact output fields used in the examples: reasoning, risk_level, justification, confidence.

Input format (what you will be given):
- assessment_question: (always the same ROB1 question about allocation sequence generation)
- ideal_evidence: a short text excerpt from a study describing (partially or fully) how participants were assigned.

Task objective:
- Decide whether the evidence supports Low, Unclear, or High risk of bias for allocation sequence generation and justify that decision concisely and explicitly with reference to the evidence.

Rules, domain knowledge, and decision algorithm (use this exactly when you assess):

1. Focus only on sequence generation (ROB1-Sequence Generation). Do NOT conflate this with allocation concealment or other biases. You may mention concealment only to note it is not assessed here.

2. Extract explicit indicators from the evidence. Look for clear method keywords. Consider these examples of adequate vs inadequate indicators:

   Adequate methods → support Low risk when explicitly described:
   - "random number table"
   - "computer-generated random sequence" / "random-number generator"
   - "coin toss" or "random draw" when described as applied to allocate participants
   - randomization list prepared in advance by an independent person/center
   - "block randomization" or "stratified randomization" with method stated (and not clearly deterministic)

   Inadequate / non-random methods → support High risk:
   - alternation, date of birth, hospital record number, order of enrolment, clinician judgement, or any systematic rule ("first come first served", "odd/even date")
   - allocation by predictable method (e.g., "assigned according to medical record number")

   Vague/insufficient information → support Unclear risk:
   - only says "randomized" or "simple randomisation" without specifying the mechanism
   - describes equal group sizes or balanced groups without providing the randomization method
   - any statement that asserts randomization but gives no method details

3. Decision flow (apply in order):
   a. If evidence explicitly states an adequate random method (see Adequate list), assign Low.
   b. Else if evidence explicitly states a non-random / quasi-random method (see Inadequate list), assign High.
   c. Else if evidence only gives vague/insufficient statements ("randomized", "simple randomisation", "participants were divided equally", or no relevant detail), assign Unclear.
   d. If evidence contains conflicting statements (e.g., calls study randomized but elsewhere describes alternation), treat as High if any clear non-random process is described; otherwise treat as Unclear.

4. Do NOT infer adequacy from indirect signs:
   - Do not infer Low just because groups are equal in size or because the study calls itself "randomized" without method.
   - Do not assume an adequate computer method unless it is named.

5. Justification content requirements:
   - Start with a concise statement of what the evidence says (quote or paraphrase the critical phrase).
   - Link that phrase to the rule that led to your risk_level (e.g., "explicit random number table → Adequate → Low").
   - If Unclear or High, name why (missing details or explicit non-random method) and what specific additional information would change the judgment (e.g., "If the authors reported a computer-generated sequence, this would support Low").
   - Keep the justification in bullet points and be explicit about which parts of the excerpt you used.

6. Reasoning:
   - Provide a short (1–3 sentence) summary explaining your interpretive logic (this is the concise chain-of-thought style explanation of how you mapped evidence → risk level). Do not include long introspective chains or extensive private reasoning—keep it outcome-focused and factual.

7. Confidence score:
   - Provide a numeric confidence between 0 and 1.
   - Guidance:
     - High confidence (0.8–0.95) when the evidence clearly and unambiguously matches the Adequate or Inadequate lists.
     - Moderate confidence (0.5–0.79) when evidence suggests but is not fully explicit.
     - Low confidence (0.2–0.49) when evidence is very sparse or ambiguous.
   - Do not use 0 or 1; avoid extremes.

8. Output format (exact fields):
   - reasoning: <concise summary, 1–3 sentences>
   - risk_level: one of {Low, Unclear, High}
   - justification: bullet-point list referencing the evidence and applying the rules above
   - confidence: numeric (0–1) per guidance

9. Tone and style:
   - Be concise, factual, and explicit.
   - Do not speculate beyond what the evidence supports; if something is unknown, call it Unclear and say what would be needed to upgrade/downgrade the judgment.

Examples to follow:
- "Randomization was performed by a study coordinator using a random number table prepared prior to recruitment." → Low (explicit random number table). High confidence (e.g., 0.8–0.9).
- "This study was designed as a randomized... trial." → Unclear (study claims randomized but no method). Moderate/low confidence (e.g., 0.5–0.65).
- "Divided into two equal groups using the simple randomisation method." → Unclear, unless the mechanism is specified. Do not infer Low based on the phrase "simple randomisation" alone.

Adhere strictly to these rules each time you assess an input.
\end{lstlisting}

\subsection{Mistral small 3.1 with Claude-3.5-sonnet}
\begin{lstlisting}[language=]
Task: Assess the risk of bias for allocation sequence generation in research studies using the ROB1 (Risk of Bias 1) criteria. Evaluate how participants were assigned to different study groups and determine the level of potential bias in the allocation process.

Input Format:
- assessment_question: Question about whether the allocation sequence was adequately generated
- ideal_evidence: Description of how participants were allocated to study groups

Expected Outputs:
1. reasoning: Analysis explaining how the evidence relates to allocation sequence generation
2. risk_level: Assessment as "Low", "High", or "Unclear" based on these criteria:
   - "Low" risk when:
     * Simple random methods like coin toss (when properly implemented and documented)
     * Computer-generated sequences
     * Random number tables
     * Dedicated randomization software
     * Clear description of stratification variables and block sizes if used
   - "High" risk when:
     * Systematic allocation methods (e.g., alternating sequence, A/B pattern)
     * Allocation based on enrollment order
     * Any predictable or non-random sequence
     * Allocation based on participant preference
   - "Unclear" risk when:
     * Only mentions "predetermined allocation list" without details
     * Only mentions "random" or "randomized" without methodological details
     * Mentions stratification variables but no details about sequence generation
     * Insufficient information about how the sequence was created

3. justification: Detailed explanation supporting the risk level, including:
   - Specific quotes or evidence from the text
   - Analysis of whether the method could introduce systematic differences
   - Impact on group comparability
   - Discussion of any stratification or blocking methods used

4. confidence: Score between 0-1 reflecting assessment certainty
   - High (>0.9): Explicit description of randomization method with technical details
   - Medium (0.7-0.9): Some methodological information but missing specific details
   - Low (≤0.7): Minimal or vague information about allocation

Key Assessment Guidelines:
1. Simple randomization methods like coin toss, when properly implemented and documented, should be considered low risk
2. Any allocation based on participant preference or non-random criteria should be considered high risk
3. Predetermined allocation lists without details about their generation should be considered unclear risk
4. The presence of a research coordinator or observer does not affect the risk level - focus on the method itself
5. Consider whether the described method could allow prediction of future assignments
6. When in doubt between risk levels, focus on the specific details of the sequence generation method
7. The mere mention of "randomization" without supporting details is insufficient for low risk

Domain-Specific Knowledge:
1. Coin toss, when properly implemented and documented, is an acceptable randomization method
2. Participant preference-based allocation always introduces high risk of bias
3. The presence of observers or coordinators does not affect the randomization quality
4. Predetermined lists require explicit details about their generation method
5. Transparency in procedure does not compensate for inadequate randomization methods

Remember: Focus specifically on how the sequence was generated, not on other aspects of randomization like allocation concealment or implementation.
\end{lstlisting}

\end{document}